%% file: paper.tex
\documentclass[letterpaper, 10 pt, conference]{ieeeconf}  

\IEEEoverridecommandlockouts                              

\input{macro.tex}
\usepackage[compress]{cite} 

\title{\LARGE \bf
\ours: A Vision-Language-Action Model for Urban Micromobility 
}

\author{Anqi Li$^{1,2,*}$\quad Zhiyong Wang$^{2,*}$\quad Jiazhao Zhang$^{1,2,*}$\quad Minghan Li$^{2}$ \\Yunpeng Qi$^{3,4}$\quad Zhibo Chen$^{3}$\quad Zhizheng Zhang$^{2,4,\dagger}$\quad He Wang$^{1,2,4,\dagger}$\\
\normalsize{$^{1}$Peking University \quad
$^{2}$Galbot \quad
$^{3}$USTC \quad 
$^{4}$BAAI  } \\
Project Page: \url{https://pku-epic.github.io/UrbanVLA-Web/}
\thanks{*Joint first authors \quad $\dagger$Corresponding authors}
}

\input{package}

\begin{document}

\let\oldtwocolumn\twocolumn
\renewcommand\twocolumn[1][]{
    \oldtwocolumn[{#1}{
        \centering
        \includegraphics[width=1\textwidth]{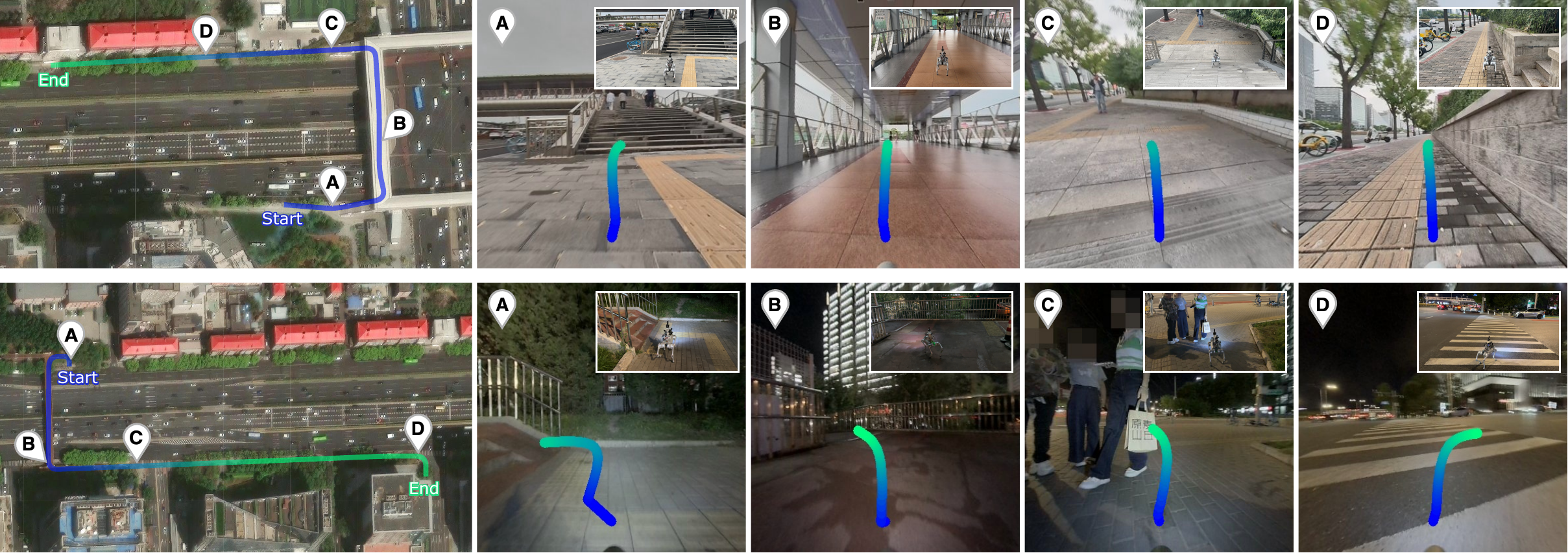} 
        \captionof{figure}{\textbf{Real-world deployments of \ours} demonstrate zero-shot generalization across diverse environments featuring unseen layouts, dynamic obstacles, and varying illumination, and highlight its ability to perform long-horizon urban micromobility tasks spanning over 500 meters.
        }      
        \vspace{5pt}
        \label{fig:teaser}
    }]
}

\maketitle



\input{section/abstract}

\input{section/intro}

\input{section/related_works}

\input{section/method}

\input{section/experiments}

\input{section/conclusion}

\bibliographystyle{IEEEtran}
\bibliography{references}


\end{document}

%% file: macro.tex
\def\ours{UrbanVLA\xspace}

\usepackage[ruled,vlined]{algorithm2e}
\DontPrintSemicolon



%% file: package.tex
\usepackage[table]{xcolor}
\usepackage{titletoc}
\usepackage{tocloft}
\usepackage{lipsum}
\usepackage{url}
\usepackage{graphicx}
\usepackage{subcaption}
\usepackage{listings}
\usepackage[nolist]{acronym}
\usepackage{amsmath}
\usepackage[font={small}]{caption}
\usepackage[export]{adjustbox}
\usepackage{amssymb}
\usepackage{bm}
\usepackage{booktabs}
\usepackage{balance}
\usepackage{color}
\usepackage{dsfont}
\usepackage{esvect}
\usepackage{float}
\usepackage{graphicx}
\usepackage{import}
\usepackage{mathtools}
\usepackage{multirow}
\usepackage{multicol}
\usepackage{flushend}
\usepackage{outline}
\usepackage{paralist}
\usepackage{siunitx}
\usepackage{stackengine}
\usepackage{stmaryrd}
\usepackage{systeme}
\usepackage{soul}
\usepackage{url}
\usepackage{tikz}
\usetikzlibrary{tikzmark}
\usetikzlibrary{calc}
\usepackage[normalem]{ulem}
\let\labelindent\relax
\usepackage{enumitem}
\usepackage{algorithm2e}
\usepackage{scalerel}
\usepackage{makecell}
\usepackage{stfloats}

\definecolor{citecolor}{HTML}{0071bc}
\definecolor{brightorange}{RGB}{255,225,0} 

\usepackage[pagebackref=true,breaklinks=true,urlcolor=blue,colorlinks,citecolor=citecolor,bookmarks=false]{hyperref}

%% file: section/abstract.tex
\begin{abstract}


Urban micromobility applications, such as delivery robots, demand reliable navigation across large-scale urban environments while following long-horizon route instructions. This task is particularly challenging due to the dynamic and unstructured nature of real-world city areas, yet most existing navigation methods remain tailored to short-scale and controllable scenarios. 
Effective urban micromobility requires two complementary levels of navigation skills: low-level capabilities such as point-goal reaching and obstacle avoidance, and high-level capabilities, such as route–visual alignment. To this end, we propose \ours, a route-conditioned Vision-Language-Action (VLA) framework designed for scalable urban navigation. Our method explicitly aligns noisy route waypoints with visual observations during execution, and subsequently plans trajectories to drive the robot.
To enable \ours to master both levels of navigation, we employ a two-stage training pipeline. The process begins with Supervised Fine-Tuning (SFT) using simulated environments and trajectories parsed from web videos. This is followed by Reinforcement Fine-Tuning (RFT) on a mixture of simulation and real-world data, which enhances the model's safety and adaptability in real-world settings.
Experiments demonstrate that \ours surpasses strong baselines by more than $55\%$ in the SocialNav task in MetaUrban. Furthermore, \ours achieves reliable real-world navigation, showcasing both scalability to large-scale urban environments and robustness against real-world uncertainties.

\end{abstract}

%% file: section/intro.tex
\section{Introduction}

Urban micromobility~\cite{abduljabbar2021role, oeschger2020micromobility}, encompassing lightweight embodied platforms such as delivery robots, assistive wheelchairs, and guide robots, is emerging as one of the most promising applications of embodied intelligence. To achieve deployment at scale, these systems must operate in dynamic and uncertain environments that, unlike the structured road networks of autonomous driving, unfold in unstructured pedestrian spaces, making reliable and scalable navigation a fundamental challenge.

Urban micromobility is achieved predominantly through visual navigation, serving as the primary paradigm for autonomous operation.
Traditional Simultaneous Localization and Mapping (SLAM)-based pipelines~\cite{cadena2016slam,wolcott2017lidar,hata2016road,schlichting2016localization} provide reliable navigation in constrained settings but depend heavily on detailed maps such as occupancy grids or HD maps. This reliance severely limits scalability in large and dynamic urban environments, where maintaining accurate maps is costly and often infeasible.
To address these limitations, learning-based approaches ~\cite{liu2025citywalker,roth2024viplanner,Yang2023iPlannerIP, zhang20233d,he2025from,zhang2025crestescalablemaplessnavigation,wu2025urbansim} formulate urban micromobility as a point-goal navigation task, with foundation model methods such as CityWalker~\cite{liu2025citywalker} leveraging waypoints from consumer navigation tools (\emph{e.g.}, Google Maps) to provide high-level guidance. However, navigation tools preserve only coarse route-level topological continuity while neglecting geometric accuracy, causing frequent misalignment between waypoints and the physical world.This makes existing point-goal navigation methods difficult to use in large-scale, real-world urban environments.



Recently, navigation-purpose VLA methods have demonstrated both strong performance and generalization~\cite{zhang2024navid,cheng2024navila,zhang2024uni,wei2025streamvln,wang2025trackvla,zhang2025embodiednavigationfoundationmodel,hirose2025omnivla}. 
By fine-tuning vision-language models, they align natural language commands with visual observations, enabling robots to ground high-level instructions into low-level actions for reliable navigation in previously unseen environments.
Despite this progress, VLAs remain inadequate for navigating urban environments over long distances. They must interpret noisy routes from navigation apps by aligning them with visual cues and commonsense knowledge. Furthermore, they need to adhere to a complex set of rules, including traffic signals, sidewalk etiquette, while simultaneously adapting in real time to dynamic obstacles, varied terrain, and dense pedestrian flows.

To address these challenges, we propose a route-conditioned Vision-Language-Action (VLA) framework for urban micromobility. Our VLA model takes structured route descriptions (roadbooks) as input and directly predicts trajectory waypoints for route following.
The VLA learns to align these "roadbooks" with visual observations, generating local trajectories that translate high-level routes into low-level navigation waypoints. This process enables reliable, long-horizon navigation over large areas, thereby allowing our framework to leverage noisy navigation tools as effective guidance for robust and scalable urban micromobility.

To train our route-conditioned VLA, we build upon a pre-trained navigation foundation model~\cite{zhang2025embodiednavigationfoundationmodel} and adopt a two-stage training approach. In the first stage, we perform supervised fine-tuning on two complementary tasks: (i) a navigation task using trajectories from the MetaUrban simulator~\cite{wu2025metaurban} and Sekai web  navigation videos~\cite{li2025sekai}, where navigation ‘roadbooks’ are derived from ground-truth paths and paired with visual observations, and (ii) a real-world Video Question Answering (VideoQA) task~\cite{shen2024longvu,li2025sekai} to enhance real-world scene understanding. In the second stage, we apply offline reinforcement fine-tuning using Implicit Q-Learning (IQL)~\cite{kostrikovoffline} on a sim-real aggregated dataset. This two-stage training equips our VLA model with the ability to interpret noisy routes, adhere to navigation norms, and adapt to real-world urban complexity.

%

Experiments demonstrate that our method achieves $55\%$ and $56\%$ performance improvement compared to baselines in the SocialNav task in \textit{MetaUrban-test} and \textit{MetaUrban-unseen} benchmark~\cite{wu2025metaurban}, respectively. Real-world experiments further prove the generalbaility of our method. We summarize our contribution as follows:



\begin{itemize}

\item We propose the first route-conditioned VLA designed for urban micromobility, which integrates the guidance of high-level urban navigation tools with vision-language policy learning.

\item We develop a simulation-to-reality pipeline, leveraging a route-lifting algorithm to formulate a sim-real aggregated dataset, achieving SOTA performance in simulation and demonstrating real-world generalization.

\item We improve safety-critical behaviors by introducing IQL-based reinforcement fine-tuning, improving obstacle avoidance, pedestrian interaction, and traffic compliance.

\end{itemize}

%% file: section/related_works.tex
\section{Related Works}

\noindent\textbf{Robot Urban Navigation.} 
Urban navigation has long relied on SLAM-based systems~\cite{cadena2016slam,zhang2014loam,wolcott2017lidar,schreiber2013laneloc}, typically using pre-built maps (\emph{e.g.}, point clouds or HD maps) for localization and planning. 
Although effective in structured settings, such methods require costly map construction and careful tuning, limiting scalability.
Learning-based alternatives reduce map dependence. OpenNav~\cite{qiao2025opennav} employs Multimodal Large Language Models (MLLMs) to generate Bird’s-Eye-View (BEV) maps and convert free-form instructions into trajectories, while CityWalker~\cite{liu2025citywalker} learns policies from large-scale urban videos. 
However, existing methods remain constrained to sensor–action correlations and lack comprehensive scene understanding. To address this, \ours introduces an end-to-end VLA framework for urban navigation that integrates multimodal perception and semantic reasoning to embed scene semantics into route-conditioned decision-making in dynamic environments.

\noindent\textbf{VLA Models for Navigation.}  
VLA models extend pre-trained VLMs with action generation and show strong generalization in embodied navigation~\cite{zhang2024navid,cheng2024navila,zhang2024uni,wei2025streamvln,wang2025trackvla,zhang2025embodiednavigationfoundationmodel,hirose2025omnivla}. 
By jointly modeling vision, language, and action, they enable instruction-following under real-time observations.  
NavFoM~\cite{zhang2025embodiednavigationfoundationmodel} trains a navigation foundation model on cross-embodiment data for consistent multi-platform performance. 
However, existing models face challenges in urban navigation, which requires sophisticated spatial reasoning, commonsense adherence, and safety-critical decision-making. To address this, we design a sim-to-real fine-tuning pipeline using the MetaUrban simulator and the Sekai dataset to acquire navigation policies for dynamic urban environments.

\noindent\textbf{Reinforcement Learning for VLA.}
Reinforcement Learning(RL) has become a key post-training strategy for VLA models~\cite{chen2025conrft,liu2025can,huang2025co,zhang2025balancing}, improving robustness over imitation learning. 
In navigation, VLN-R1~\cite{qi2025vln} applies temporally-decayed rewards for long-horizon optimization, and OctoNav~\cite{gao2025octonav} uses hierarchical rewards with online adaptation. 
However, VLA models for urban navigation face safety-critical challenges such as pedestrian interactions and traffic rule compliance. To address this, we use Implicit Q-Learning (IQL)~\cite{kostrikovoffline} for reinforcement fine-tuning to enhance safety-aware decision-making in dynamic urban environments.

%% file: section/method.tex
\section{Method}

\subsection{Problem Formulation}

\label{subsec:probform}

\noindent\textbf{Task Definition. }We formulate the route-conditioned urban navigation task as follows: at the current time step $T$, given a high-level target route formulated by a sequence of 2D coordinates $\mathcal{R} =  \{r_0, ..., r_n\}$ where $r_i \in \mathbb{R}^2$ is a planar coordinate sampled from the target route under the agent's egocentric frame, and a sequence of RGB image observations $\mathcal{O}_{\text{vis}} = \mathbf{o}_{1:T}^{1:C} \in \mathbb{R}^{W \times H \times 3}$ taken by $C$ different cameras at time steps $\{1,...,T\}$, 
the agent is required to learn a navigation policy $\pi (\mathcal{R}, \mathcal{O}_{\text{vis}}) \mapsto \tau$, where $\tau = \{\tau_1, \tau_2, ...\tau_N\}$ is a navigation trajectory formulated by $N$ waypoints with $\tau_i \in \mathrm{SE}(2)$ representing the predicted 2D position and orientation under the current egocentric frame that safely drives the agent along the target route toward its destination.

\input{figure/pipeline}

\noindent\textbf{Pipeline Overview. }\label{subtit: pipoverview}Figure~\ref{fig:pipeline} shows the overall pipeline of our approach. We leverage a pre-trained navigation foundation model NavFoM~\cite{zhang2025embodiednavigationfoundationmodel} as our base model, and apply a two-stage fine-tuning strategy via supervised fine-tuning (SFT) and reinforcement fine-tuning (RFT), respectively. Specifically, we apply a prompt template to encode the high-level ‘roadbook’ instructions
into a linguistic form $\mathcal{I}$. Following existing VLM approaches~\cite{liu2023llava, li2023llama, shen2024longvu}, we embed $\mathcal{I}$ to obtain language tokens $E_L$, as well as encode visual observations $\mathcal{O}_{\text{vis}}$ using pre-trained vision encoders to obtain visual tokens $E_{1:T}^{1:C}$. Together, we feed $E_L$ and $E_{1:T}^{1:C}$ into the Large Language Model (LLM) backbone. In the SFT stage, following previous work~\cite{wang2025trackvla}, the dual-branched VLA learns to perform two types of tasks, VideoQA and route-conditioned navigation. We decode the tokens generated using a language head and an action head, respectively, to acquire the linguistic answer and navigation trajectory.



In the RFT stage, we further fine-tune \ours on a hybrid dataset that combines expert demonstrations collected from both simulation and real-world environments.
We adopt Implicit Q-Learning (IQL) ~\cite{kostrikovoffline}, a widely used offline RL algorithm, to effectively utilize these limited hybrid data while mitigating overoptimism in out-of-distribution (OOD) samples.
To estimate the action-state function $Q$ and value function $V$ in IQL
that take current state as input, 
we encode language instruction $\mathcal{I}$ and visual observation $\mathcal{O}_{\text{vis}}$ into a unified state representation $s$ using the well-tuned vision-language backbone after the SFT stage. 
The reward function $r(s, a)$ is delicately designed, considering both trajectory efficiency and navigation safety, to allow efficient data collection in the real world, as well as alignment between simulation and real world. 


\subsection{\ours Architecture}
\label{subtit:routeencode}
\noindent\textbf{High-Level Route Encoding. }
The high-level route instructions in the urban navigation task should be converted into a form that is interpretable by VLA models and aligned with the data schema of the prevalent urban navigation tools~\cite{amap2023} to facilitate large-scale deployment. 
%
Consequently, we convert the route instructions into a structured linguistic representation comprising two components. First, a set of waypoints sampled from the high-level route provides the agent with the overall geometry and orientation of the forthcoming path. Second, distance and direction instructions for the next turn (for example, ‘turn right in 30 meters’) convey essential information to transition between blocks, a critical scenario for successful urban navigation.


Specifically, given a high-level navigation route $\mathcal{R}$, we first resample the upcoming $D$ meters of the route trajectory at a distance of $d$ meters (we set $D = 40$ and $ d=2$, resulting in 20 waypoints), and convert them into the robot frame. Subsequently, when training, we apply a corner detection algorithm (see Section~\ref{subsec:impdetail}) to segment the route into blocks and then derive block-level distance and direction cues from these segments; while in a real-world setting, this information can be directly acquired from the API of the city navigation tool. Finally, we formulate the above information into an instruction template to obtain the navigation instruction $\mathcal{I}$. 


\noindent\textbf{VLA Model Forwarding. }Given multi-view RGB observations $\mathcal{O}_{\text{vis}} = \mathbf{o}_{1:T}^{1:C} \in \mathbb{R}^{W \times H \times 3}$, for route-conditioned urban navigation tasks, we apply a visual sliding window to retain the nearest $k$ frames $\mathcal{O_{\text{retain}}} = \mathbf{o}_{T-k+1:T}^{1:C}$. Following recent advanced VLM works~\cite{tong2024cambrian1, shen2024longvu, kim2024openvla}, we encode visual information using two pre-trained vision encoders (DINOv2~\cite{oquab2023dinov2} and SigLIP~\cite{zhai2023siglip}) and concatenate the visual features obtained in the channel dimension to formulate the final visual features $v^{1:C}_{1}, ... v^{1:C}_{T-1}, v^{1:C}_T$. Subsequently, we down-sample the features with grid pooling strategy, and use a cross-modal projector (double-layer MLP)~\cite{liu2023llava} to project the visual features into the embedding space of the LLM backbone and acquire the visual tokens $E_{1:T}^{1:C}$. We then embed the navigation instruction $\mathcal{I}$ in language tokens $E_L$. 
Together, we feed all tokens into an LLM backbone (Qwen2~\cite{qwen2}). The model generates tokens in two manners: for navigation tasks, we capture the generated action token $E_T^A$ at the current time step and decode it through an MLP-based action model to obtain the navigation trajectory $\tau$: 
\begin{equation}
\begin{split}
\label{alg: action decode}
    E_T^A &= \text{LLM}(E_L, E_{1:T}^{1:C}), \\
    \tau &= \text{ActionModel}(E_T^A); 
\end{split}
\end{equation}
while for VideoQA tasks, the model autogressively generates a set of language tokens, which are then decoded through the language model head, as shown in Figure~\ref{fig:pipeline}. 


\subsection{Training Strategy}


\noindent\textbf{Supervised Fine-Tuning. }We first apply Supervised Fine-tuning (SFT) to the base model NavFoM~\cite{zhang2025embodiednavigationfoundationmodel}. In this stage, the model learns from urban navigation demonstrations generated by a PPO expert in simulation~\cite{wu2025metaurban}, as well as web-scale urban travel data that capture human navigation behavior in real-world environments, using a Mean Squared Error (MSE) loss. The SFT stage is designed to instill basic goal-reaching capabilities while exposing the model to the diversity and complexity of urban navigation tasks, thus enhancing its generalization to real-world scenarios.




A key challenge in leveraging such demonstration data is 
the lack of ‘roadbook’ instructions.
Real-world demonstrations typically provide only the ground-truth trajectory, while simulators often offer perfect route information generated by collision-free planning algorithms such as ORCA~\cite{van2011reciprocal} and P\&R~\cite{Wilde2014PushAR}. Using such idealized routes directly as conditions may cause the model to overfit to the input trajectory, thus compromising its generalizability in real-world scenarios.

To address this problem, we introduce \textit{Heuristic Trajectory Lifting (HTL)}, a heuristic algorithm that lifts high-level route information from the raw trajectory in urban navigation data, encouraging the model to learn from visual cues rather than relying solely on idealized route inputs. The raw trajectory is first preprocessed: a Savitzky–Golay filter~\cite{savitzky1964smoothing} is applied to denoise web trajectories, while simulator-generated smooth trajectories are used directly. The self-intersecting or otherwise low-quality paths are then removed. Next, significant turning points are detected (see Section~\ref{subsec:impdetail}) to 
split the trajectory into several segments. 
To capture the ambiguity of real-world navigation, each segment is perturbed with Gaussian positional noise, reflecting that high-level directives (\emph{e.g.}, ‘go straight’) correspond to a corridor of feasible paths rather than a single curve. Finally, noisy segments are smoothly merged and resampled in a fixed spatial step, resulting in the abstracted route $\mathcal{R}$. 
In summary, this pipeline allows us to generate a large-scale, unified dataset for the SFT stage
from both simulated and real-world sources with automatically anotated ‘roadbook’ instructions.

\noindent\textbf{Reinforcement Fine-Tuning. }Building on the capabilities acquired during SFT, \ours demonstrates strong performance in
navigating diverse urban environments, such as intersections, turns, and varying street layouts. 
To further improve its skills, particularly in collision avoidance and handling ambiguous cues, we adopt an offline RL approach based on Implicit Q-Learning (IQL) ~\cite{kostrikovoffline}, which suits well for offline data and mitigates out-of-distribution action issues. 

We formulate the route-conditioned urban navigation task as a Partially Observable Markov Decision Process (POMDP) $\mathcal{M} := (\mathcal{S}, \mathcal{A}, \mathcal{O}, \mathcal{T}, r, \gamma)$, 
where $\mathcal{S}, \mathcal{A}, \mathcal{O}, \mathcal{T}, r$, and $\gamma$ denote the state space, action space, observation space, transition model, reward function, and discount factor, respectively.
%
For IQL, we first train value networks $Q_\theta(s,a)$ and $V_\psi(s)$ for advantage estimation. To properly represent the input state $s \in \mathcal{S}$, we employ the VLA model after the SFT stage (denoted as \ours-SFT) as the encoder. Specifically, at current timestep $T$, the observation $o_T = (\mathcal{O}_\text{vis}, \mathcal{I}) \in \mathcal{O}$ is fed into \ours-SFT, and the hidden state from the $n$-th transformer layer of the LLM backbone is extracted to serve as the state embedding $s = H_{T}^{(n)} \in \mathbb{R}^{\text{dim}}$ (as shown in Figure~\ref{fig:pipeline}), where $\text{dim}$ is the hidden dimension of the LLM.
We empirically set $n=17$, as we observe that mid-layer hidden states yield more robust value estimation than top-layer states. This is potentially due to the latter being overly specialized for predicting action logits. We find this representation effectively encodes the visual and language context, offering task-aware embeddings that stabilize Q-learning ~\cite{huang2025co}. For the input action, we simply vectorize the predicted trajectory: $a := \text{vec}(\tau) \in \mathbb{R}^{3N}$, with $\tau$ and $N$ defined in Section~\ref{subsec:probform}. 

Based on the above formulation, IQL learns a value function $V_\psi(s)$ and a Q-function $Q_\theta(s,a)$ from the offline dataset $\mathcal{D}$, and the policy $\pi (s)$ is updated via an advantage-weighted regression (AWR) objective
\begin{equation}
    \label{eq:iql_loss}
\mathcal{L}^\text{IQL}_\pi =
\mathbb{E}_{(s,a)\sim\mathcal{D}}
\Big[
\exp\big(\beta A(s,a)\big)
||a - \pi (s)||^2
\Big],
\end{equation}
where $A(s,a)=Q_\theta(s,a)-V_\psi(s)$ is the advantage estimation, and $\beta$ is an inverse temperature parameter that trades between imitation and performance improvement~\cite{kostrikovoffline}. 




The reward function $r(s, a)$ is designed for: (i) ease of acquisition during teleoperation to minimize post-processing, and (ii) sim-real consistency to provide a unified learning objective and enhance data efficiency. We formulate the reward function as 
\begin{equation}
r(s,a) =
\lambda_{\text{comp}}\, l_{\text{completion}}
- \lambda_{\text{coll}}\, \mathds{1}_{\text{collision}}
- \lambda_{\text{dev}}\, \mathds{1}_{\text{deviation}},
\label{eq:reward}
\end{equation}
where $l_{\text{completion}}$ denotes the increment of trajectory completion aligned with the ground-truth route, while $\mathds{1}_{\text{collision}}$ and $\mathds{1}_{\text{deviation}}$ indicate whether a collision or an excessive deviation from the route corridor occurs, respectively. 
The weighting coefficients $\lambda_{\text{comp}}, \lambda_{\text{coll}}, \lambda_{\text{dev}}$
are adjusted to balance efficiency and safety and are specified in Section~\ref{subsec:impdetail}. 
This formulation relies only on readily available or easily measurable quantities (see the user interface in Figure~\ref{fig:setup}), and enables the seamless transfer of the learned policy to real environments. 

Based on this design, we collect a sim-real aggregated dataset for the RFT stage, comprising 2,400 episodes (approximately 40 hours) in the MetaUrban simulator using a PPO expert, along with roughly 8 hours of real-world demonstrations via human teleoperation. Please refer to Section~\ref{subsec:impdetail} for details. 
In summary, during the RFT stage, the large scale simulation data facilitate rapid convergence of the Q-learning, and the teleoperation data is efficiently learnt to strengthen the model's real-world safety performance and corner case decisions.



\noindent\subsection{Implementation Details}
\label{subsec:impdetail}

Our model is trained on a cluster server equipped with 8 NVIDIA H100 GPUs for approximately 12 hours, resulting in 96 GPU hours in total. The VideoQA dataset is collected from LongVU~\cite{shen2024longvu} and Sekai~\cite{li2025sekai}. Unlike the sliding-window mechanism introduced in the navigation task, we retain all visual frames before feeding them into the model. We supervise the result using cross-entropy loss. 

For the corner detection algorithm mentioned in Section~\ref{subtit:routeencode}, specifically, we adopt a window-based detection algorithm: for each point, we compute the turning angle between the vectors formed by its neighbors in a window of $k$ points. Points with angles above a threshold are marked as candidates. The subsequent candidates are merged by taking the middle point, and a greedy selection step enforces a minimum arc-length spacing to remove redundant corners.

In the RFT stage, the real-world data was collected in the city area of Beijing. For this dataset, we collected visual observations, navigation route instructions obtained by querying navigation tool APIs in real time, and reward signals either annotated through the user interface or generated using records from a real-world LiDAR-Odometry system. We intentionally collected several scenes in which the navigation information is inconsistent with real-world conditions. Notably, a small amount of sub-optimal real-world data (that yields negative reward terms) was also intentionally collected to ensure the stability of Q-learning ~\cite{kostrikovoffline, kumar2020conservative}. The weighting coefficients $\lambda_{\text{comp}}$, $\lambda_{\text{coll}}$, and $\lambda_{\text{dev}}$ in the reward function are set to $0.5$, $1$, and $1$, respectively.



%% file: figure/pipeline.tex
\begin{figure*}[t]
  \centering
  \includegraphics[width=0.85\textwidth]{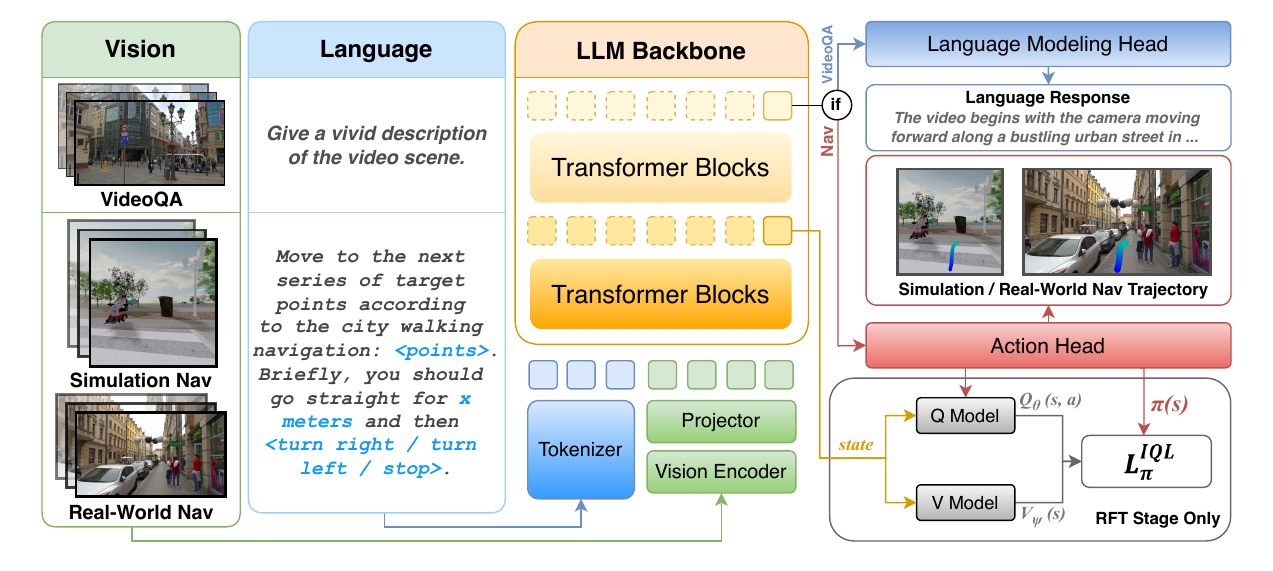}
  \caption{\textbf{Overview of \ours.} We collect diversified VideoQA data and urban micromobility demonstrations to train the model via a two-stage pipeline. In the SFT stage, \ours learns essential urban navigation capabilities such as goal-reaching, collision avoidance, and social compliance; in the RFT stage, we refine the model using a sim-real aggregated dataset with IQL, enhancing robustness in real-world scenarios. }
  \label{fig:pipeline}
\end{figure*}

%% file: section/experiments.tex
\section{Experiments}

\subsection{Experiment Setup}
\label{subsec: experiment setup}

\noindent\textbf{Simulation Setup. }We conduct our experiments based on the Point Navigation (PointNav) and Social Navigation (SocialNav) benchmarks built on the MetaUrban simulator~\cite{wu2025metaurban}. These two tasks provide an insightful examination on our model's capabilities of route following, collision avoidance, social compliance, and generalizability to diverse scenery layout. To allow fair comparison, our model is trained on 12,800 scenes in the \textit{MetaUrban-train} dataset, and is tested on 1,000 scenes in the \textit{MetaUrban-test} and 100 scenes in the \textit{MetaUrban-unseen}. 

In our evaluation, we choose a wheelchair embodiment. While the baselines output continuous steering and acceleration~\cite{wu2025metaurban}, \ours focuses on high-level spatial planning by predicting 2D trajectory waypoints. To execute this, the agent is directly displaced to the predicted waypoints. We bound the maximum step distance to the radius of the embodiment (1.5 meters) to ensure accurate collision detection. 


\input{figure/setup}

\noindent\textbf{Real-World Setup. }We conduct extensive real-world experiments across multiple urban blocks, covering diverse scenarios such as overpasses, pedestrian crossings, and environments with dense static and dynamic obstacles. Following NavFoM ~\cite{zhang2025embodiednavigationfoundationmodel}, our model runs on a remote server equipped with an NVIDIA RTX 4090 GPU, communicating with a Unitree Go2 quadruped via Web-ADK. The robot collects visual observations from four cameras mounted on its front, left, right, and rear, and relies on an off-the-shelf local planner to execute the trajectories generated by our model. High-level route instructions are obtained from Amap ~\cite{amap2023} via its Web API, while a GNSS module provides the real-time position and orientation of the quadruped to convert these instructions into the robot's local frame. Figure~\ref{fig:setup} illustrates our real-world deployment setup. The system operates at 2 Hz, which is sufficient for smooth and responsive navigation. During the experiments, a human safety operator monitored the model in real time and manually intervened to halt the system whenever unsafe behaviors occurred.

\input{table/benchmark}
\noindent\textbf{Baselines and Metrics. }We select the established RL/IL models from MetaUrban~\cite{wu2025metaurban} as our primary baselines, as they represent the standard state-of-the-art on the two benchmarks. We evaluate the models using the following metrics: Success Rate (SR) and Success weighted by Path Length (SPL)~\cite{anderson2018evaluation} to evaluate both the effectiveness and efficiency of the agent’s navigation; the Cumulative Cost (Cost)~\cite{li2022metadrive} evaluates the agent’s ability to avoid collisions, while the Social Navigation Score (SNS)~\cite{deitke2022retrospectives} quantifies the model’s compliance with social navigation standards.



\subsection{Quantitative Experiments}

As shown in Table~\ref{tab:benchmark}, our method significantly outperforms all baseline methods on both PointNav and SocialNav tasks in terms of SR and SPL/SNS. Specifically, our model achieves 94\% SR and 0.91 SPL on the PointNav test set, and 97\% SR and 0.95 SPL on the unseen set, surpassing baselines by more than 25\% in SR while demonstrating strong generalization capability in unseen urban environments. The superior performance in unseen scenaries with unusual object placement and sidewalk layout indicates that our approach captures systematic structural and geometric priors of urban navigation, enabling highly effective and efficient goal-reaching behavior.

In the SocialNav task, \ours attains a high Social Navigation Score (SNS) of 0.87 on the test set and 0.85 on the unseen set, significantly exceeding all LiDAR-based baselines. This suggests that our model not only avoids collisions effectively, but also adheres to social norms, such as maintaining comfortable distances and yielding to pedestrians, even though relying solely on RGB inputs.

Although our method yields a relatively higher Cost compared to some baselines (\emph{e.g.} PPO-Lag), this is reasonable given the substantially higher success rates, which indicate longer traveling distance and higher probability to encounter obstacles. Moreover, learning obstacle avoidance using only the RGB input is considerably more challenging than using LiDAR. Thus, achieving such high success rates while maintaining a bounded cost reflects a strong inherent capability to avoid obstacles and navigate socially.

In summary, our method exhibits superior performance in navigation efficiency, success rate, generalization, and social compliance, validating its effectiveness for complex urban navigation tasks.

\input{figure/gallery}

\subsection{Qualitative Results}

We evaluate \ours in several representative and challenging real-world scenarios, including overpass crossing, pedestrian interaction, street turning, and obstacle avoidance on trajectories greater than 500 meters. As illustrated in Figure~\ref{fig:gallery}, the results highlight \ours’s ability to operate in large-scale, dynamic, and unstructured urban environments. In outdoor scenes, the system is able to produce stable navigation trajectories and follow designated routes while adapting to variations in illumination, weather, and even nighttime conditions. Moreover, the examples suggest that \ours can effectively align high-level navigation instructions with visual observations, enabling correct turns at intersections, successful navigation across overpasses, and adaptation to diverse route structures. In addition, we observe that \ours is capable of avoiding static and dynamic obstacles in these scenarios, allowing the maintenance of reasonable distances to pedestrians, and bypassing unexpected objects. Together, these qualitative results indicate that \ours can integrate high-level route priors with perception to achieve reliable and socially compliant navigation in complex urban settings in the real world.

\input{table/ablation_htl}

\subsection{Ablation Study}

\textbf{Effectiveness of HTL algorithm. }We evaluate the effectiveness of the HTL algorithm in both simulation and real world settings. In simulation, we evaluate using the SocialNav test set. For real-world evaluation, we design a 
60 meters 
C-shaped sidewalk (as demonstrated in Figure~\ref{fig:gallery}) with a width of nearly 8 meters, where the robot must gradually turn right and then proceed straight. The route information from the navigation tools represents this curve roughly as a polyline and exhibits an error of approximately 10 meters from the center line of the route in the real world. We use Route Completion (RC)~\cite{zhou2024mattersenhancetrafficrule} to evaluate real-world performance, which refers to the proportion of the completed route out of the whole route, and we consider the scenario to terminate when the agent deviates from the route corridor (\emph{i.e.} crashes into non-walkable area aside the route). 

The results in Table~\ref{tab:HTL} show that ablating HTL leads to a slight improvement in simulation; 
however, RC drops dramatically in real-world experiments. We observe that most failure cases are caused by the agent repeatedly attempting to reach the shifted goal point, resulting in route deviation. 
This suggests that without HTL, \ours tends to overfit the route instructions, significantly reducing its robustness to noisy route information in real-world scenarios and limiting its scalability.

\input{table/ablation_rl}
\textbf{Effectiveness of Reinforcement Learning.} We evaluate \ours under two training strategies, SFT alone and SFT combined with RFT, on the SocialNav benchmark of \textit{MetaUrban-test} and \textit{MetaUrban-unseen}. The results in Table~\ref{tab:RFT} show that after the RFT stage, \ours consistently outperforms the SFT-only model, with improved SR and reduced Cost. In particular, the performance gain is larger on the unseen benchmark, with a $+6\%$ increase in SR and a $-0.16$ decrease in Cost, indicating the enhanced safety and generalizability of \ours after the RFT stage. 
We attribute this improvement to the safety-centric reward design, and the increased diversity introduced by real-world teleoperation data.

%% file: figure/setup.tex
\begin{figure}[t]  
  \centering
  \includegraphics[width=0.7\linewidth]{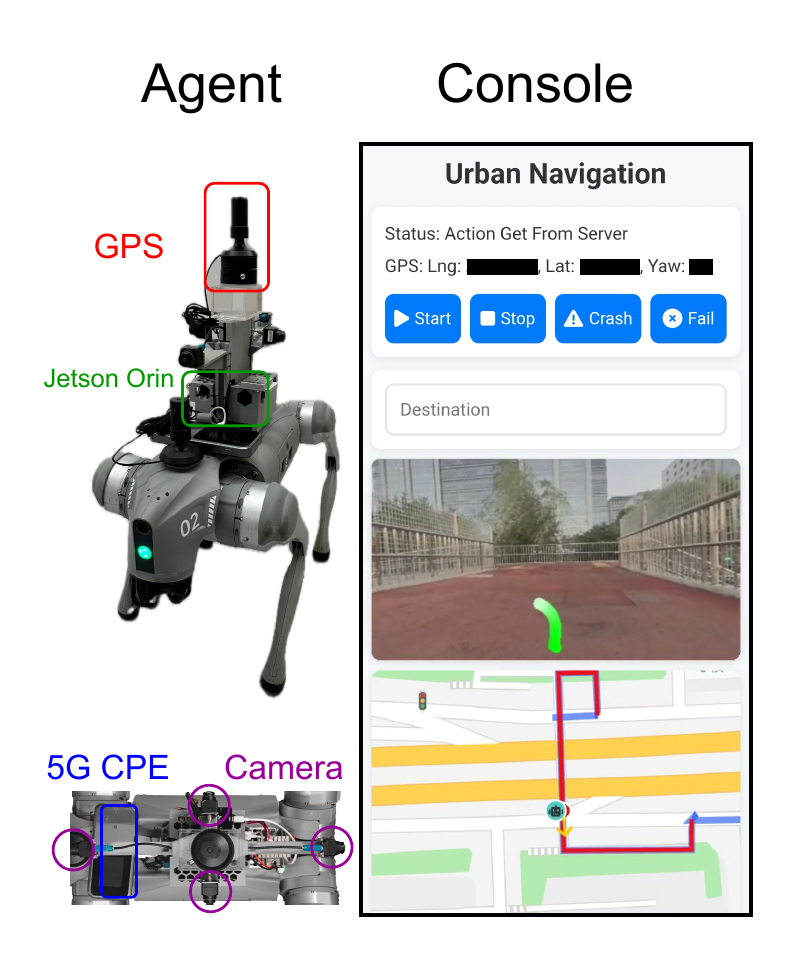} 
  \caption{\textbf{Real-world deployment of \ours. }Our system consists of a quadruped robot equipped with GPS, Wi-Fi, a camera, and an onboard computing unit, along with a mobile-deployable console for real-time monitoring, sending navigation targets, visualizing maps and model predictions, and annotating teleoperation data used for reinforcement learning. 
  }
  \vspace{-5mm}
  \label{fig:setup}
\end{figure}

%% file: table/benchmark.tex
\begin{table*}[htbp]
\footnotesize
\begin{center}
\resizebox{2\columnwidth}{!}{
    \begin{tabular}{c|c|ccc|ccc|ccc|ccc}
        \toprule
        \multirow{3}{*}{\textbf{Method}} & \multirow{3}{*}{\textbf{Observation}} & \multicolumn{6}{c|}{\textbf{PointNav}} & \multicolumn{6}{c}{\textbf{SocialNav}} \\
        \cmidrule{3-14}
        & & \multicolumn{3}{c|}{Test} & \multicolumn{3}{c|}{Unseen} & \multicolumn{3}{c|}{Test} & \multicolumn{3}{c}{Unseen} \\
        \cmidrule{3-14}
        & & \makecell{SR$\uparrow$} & \makecell{SPL$\uparrow$} & \makecell{Cost$\downarrow$} & \makecell{SR$\uparrow$} & \makecell{SPL$\uparrow$} & \makecell{Cost$\downarrow$} & \makecell{SR$\uparrow$} & \makecell{SNS$\uparrow$} & \makecell{Cost$\downarrow$} & \makecell{SR$\uparrow$} & \makecell{SNS$\uparrow$} & \makecell{Cost$\downarrow$} \\
        \toprule
        PPO\cite{schulman2017proximalpolicyoptimizationalgorithms} & LiDAR & \underline{66\%} & \underline{0.64} & 0.51 & 49\% & 0.45 & 0.78 & 34\% & 0.64 & 0.66 & 24\% & 0.57 & \underline{0.51} \\
        PPO-Lag\cite{fujimoto2019benchmarkingbatchdeepreinforcement} & LiDAR & 60\% & 0.58 & \textbf{0.41} & \underline{60\%} & \underline{0.57} & \textbf{0.53} & 17\% & 0.51 & \underline{0.33} & 8\% & 0.47 & \textbf{0.50} \\
        PPO-ET\cite{sun2021safeexplorationsolvingearly} & LiDAR & 57\% & 0.53 & \underline{0.47} & 53\% & 0.49 & 0.65 & 5\% & 0.52 & \textbf{0.26} & 2\% & 0.50 & 0.62 \\
        IQL\cite{kostrikovoffline} & LiDAR & 36\% & 0.33 & 0.49 & 30\% & 0.27 & \underline{0.63} & \underline{36\%} & \underline{0.67} & 0.39 & 27\% & 0.62 & 3.05 \\
        TD3+BC\cite{fujimoto2021minimalistapproachofflinereinforcement} & LiDAR & 29\% & 0.28 & 0.77 & 20\% & 0.20 & 1.16 & 26\% & 0.61 & 0.62 & \underline{32\%} & \underline{0.64} & 1.53 \\
        BC\cite{Bain1995AFF} & LiDAR & 36\% & 0.28 & 0.83 & 32\% & 0.26 & 1.15 & 28\% & 0.56 & 1.23 & 18\% & 0.54 & 0.58 \\
        GAIL\cite{ho2016generativeadversarialimitationlearning} & LiDAR & 47\% & 0.36 & 1.05 & 40\% & 0.32 & 1.46 & 34\% & 0.63 & 0.71 & 28\% & 0.61 & 0.67 \\
        \rowcolor{gray!15}
        \textbf{Ours} & RGB & \textbf{94\%} & \textbf{0.91} & 0.94 & \textbf{97\%} & \textbf{0.95} & 0.84 & \textbf{91\%} & \textbf{0.87} & 0.81 & \textbf{88\%} & \textbf{0.85} & 0.82 \\
        \bottomrule
    \end{tabular}
}
\end{center}

\caption{\textbf{Benchmarks.} The benchmark of PointNav and SocialNav tasks on the MetaUrban-12K dataset. We compare our method with seven strong baselines with LiDAR observation. The \textbf{best} and the \underline{second best} results are denoted by \textbf{bold} and \underline{underline}, respectively. }
\label{tab:benchmark}
\end{table*}


%% file: figure/gallery.tex
\begin{figure*}[htbp]  
  \centering
  \includegraphics[width=\textwidth]{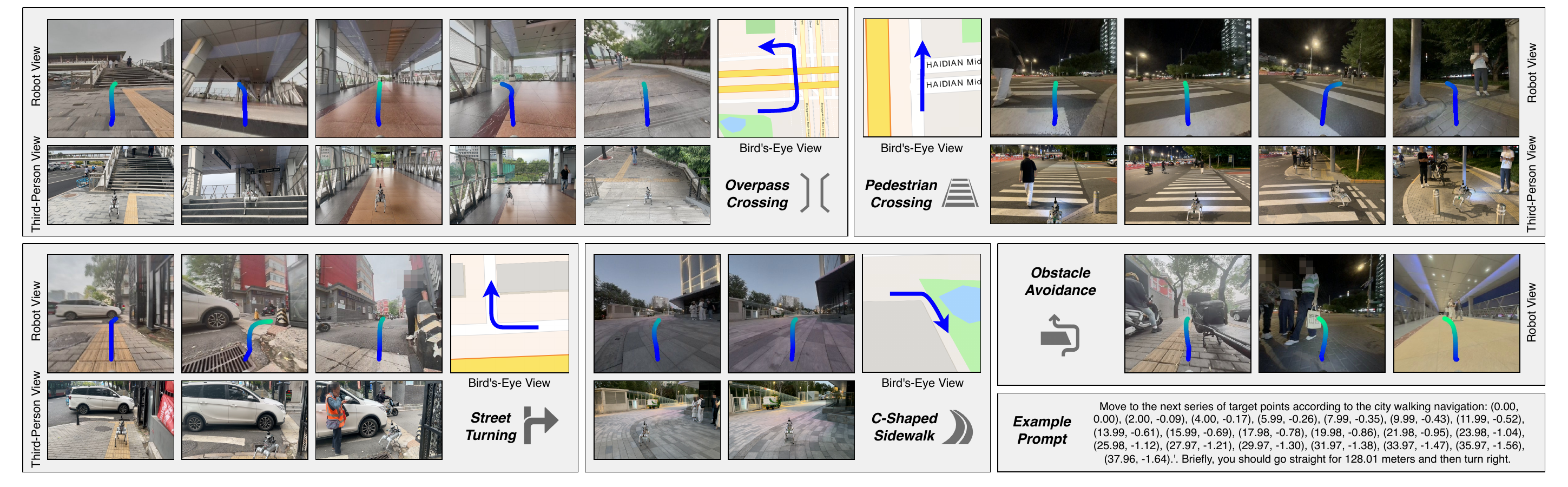} 
  \caption{\textbf{Visualization of qualitative experiment results in real-world scenarios. }We show four critical scenarios in real-world evaluation of \ours: overpass crossing, street turning, pedestrian crossing, and obstacle avoidance. \ours generates executive and reasonable trajectories shown by the blue trajectories plotted in the first person view (FPV). }
  \label{fig:gallery} 
\end{figure*}

%% file: table/ablation_htl.tex
\begin{table}[tbp]
\centering
\footnotesize
\setlength{\tabcolsep}{10pt} 
\begin{tabular}{lccc}
\toprule
\textbf{HTL Configuration} & \multicolumn{2}{c}{\textbf{Test}} & \textbf{Real World} \\
\cmidrule(lr){2-3} \cmidrule(lr){4-4}
 & \textbf{SR$\uparrow$} & \textbf{Cost$\downarrow$} & \textbf{RC$\uparrow$} \\
\midrule
Ours-SFT w/o HTL & \textbf{93\%} & \textbf{0.81} & 42\% \\
\rowcolor{gray!15}
\textbf{Ours-SFT w/ HTL} & 89\% & 0.86 & \textbf{100\%} \\
\bottomrule
\end{tabular}
\caption{\textbf{Comparison of HTL configurations.} We compare the performance of \ours after the SFT stage (denoted as Ours-SFT), with or without HTL. For simulation results, we report the model's performance on SocialNav in MetaUrban benchmark. For real-world, we report results on the C-shaped sidewalk, averaged over ten trials.
}
\label{tab:HTL}
\end{table}

%% file: table/ablation_rl.tex
\begin{table}[tbp]
\centering
\begin{tabular}{l cc cc}
\toprule
\makecell{\textbf{Training stage}} & \multicolumn{2}{c}{\textbf{Test}} & \multicolumn{2}{c}{\textbf{Unseen}} \\
\cmidrule(lr){2-3} \cmidrule(lr){4-5}
 & \textbf{SR$\uparrow$} & \textbf{Cost$\downarrow$} & \textbf{SR$\uparrow$} & \textbf{Cost$\downarrow$} \\
\midrule
Ours-SFT & 89\% & 0.86 & 82\% & 0.98\\
\rowcolor{gray!15}
\textbf{Ours} & \textbf{91\%} & \textbf{0.80} & \textbf{88\%} & \textbf{0.82} \\
\bottomrule
\end{tabular}
\caption{\textbf{Comparison across two training stages. }We compare the result of \ours before and after the RFT stage on SocialNav in MetaUrban benchmark. }
\label{tab:RFT}
\end{table}

%% file: section/conclusion.tex
\section{Conclusion}
We present a route-conditioned vision-language-action framework \ours for urban micromobility, which integrates the noisy navigation-tool routes with on-board vision to enable scalable and reliable long-horizon navigation. The model is trained through SFT on simulation-based and web video–parsed trajectories, followed by RFT with a sim-real aggregated dataset. 
The proposed approach not only improves obstacle avoidance and social compliance, but also establishes a practical framework for the deployment of embodied agents in 
open urban areas.
Future work will explore broader multimodal cues and further improve adaptability to diverse and dynamic urban contexts.

%% file: references.bib
@String(ICLR = {Int. Conf. Learn. Represent.})

@String(ICLR  = {ICLR})

@inproceedings{liu2025citywalker,
  title={Citywalker: Learning embodied urban navigation from web-scale videos},
  author={Liu, Xinhao and Li, Jintong and Jiang, Yicheng and Sujay, Niranjan and Yang, Zhicheng and Zhang, Juexiao and Abanes, John and Zhang, Jing and Feng, Chen},
  booktitle={Proceedings of the Computer Vision and Pattern Recognition Conference},
  pages={6875--6885},
  year={2025}
}

@inproceedings{qiao2025opennav,
  title={Open-nav: Exploring zero-shot vision-and-language navigation in continuous environment with open-source llms},
  author={Qiao, Yanyuan and Lyu, Wenqi and Wang, Hui and Wang, Zixu and Li, Zerui and Zhang, Yuan and Tan, Mingkui and Wu, Qi},
  booktitle={2025 IEEE International Conference on Robotics and Automation (ICRA)},
  pages={6710--6717},
  year={2025},
  organization={IEEE}
}

@inproceedings{zhang20233d,
  title={3d-aware object goal navigation via simultaneous exploration and identification},
  author={Zhang, Jiazhao and Dai, Liu and Meng, Fanpeng and Fan, Qingnan and Chen, Xuelin and Xu, Kai and Wang, He},
  booktitle={Proceedings of the IEEE/CVF Conference on Computer Vision and Pattern Recognition},
  pages={6672--6682},
  year={2023}
}

@article{wang2025trackvla,
    author  = {Wang, Shaoan and Zhang, Jiazhao and Li, Minghan and Liu, Jiahang and Li, Anqi and Wu, Kui and Zhong, Fangwei and Yu, Junzhi and Zhang, Zhizheng and Wang, He},
    title   = {TrackVLA: Embodied Visual Tracking in the Wild},
    journal = {arXiv pre-print},
    year    = {2025}
}

@article{wei2025streamvln,
  title={StreamVLN: Streaming Vision-and-Language Navigation via SlowFast Context Modeling},
  author={Wei, Meng and Wan, Chenyang and Yu, Xiqian and Wang, Tai and Yang, Yuqiang and Mao, Xiaohan and Zhu, Chenming and Cai, Wenzhe and Wang, Hanqing and Chen, Yilun and others},
  journal={arXiv preprint arXiv:2507.05240},
  year={2025}
}

@article{li2023llama,
  title={LLaMA-VID: An Image is Worth 2 Tokens in Large Language Models},
  author={Li, Yanwei and Wang, Chengyao and Jia, Jiaya},
  journal={arXiv preprint arXiv:2311.17043},
  year={2023}
}

@String(NeurIPS= {Adv. Neural Inform. Process. Syst.})

@String(NeurIPS  = {NeurIPS})

@article{anderson2018evaluation,
  title={On evaluation of embodied navigation agents},
  author={Anderson, Peter and Chang, Angel and Chaplot, Devendra Singh and Dosovitskiy, Alexey and Gupta, Saurabh and Koltun, Vladlen and Kosecka, Jana and Malik, Jitendra and Mottaghi, Roozbeh and Savva, Manolis and others},
  journal={arXiv preprint arXiv:1807.06757},
  year={2018}
}

@article{zhang2024navid,
  title={Navid: Video-based vlm plans the next step for vision-and-language navigation},
  author={Zhang, Jiazhao and Wang, Kunyu and Xu, Rongtao and Zhou, Gengze and Hong, Yicong and Fang, Xiaomeng and Wu, Qi and Zhang, Zhizheng and Wang, He},
  journal={Robotics: Science and Systems},
  year={2024}
}

@inproceedings{cheng2024navila,
title = {NaVILA: Legged Robot Vision-Language-Action Model for Navigation},
    author = {Cheng, An-Chieh and Ji, Yandong and Yang, Zhaojing and Zou, Xueyan and Kautz, Jan and Biyik, Erdem and Yin,
    Hongxu and Liu, Sifei and Wang, Xiaolong},
    booktitle = {RSS},
    year = {2025},
}

@article{zhang2024uni,
    title={Uni-NaVid: A Video-based Vision-Language-Action Model for Unifying Embodied Navigation Tasks},
    author={Zhang, Jiazhao and Wang, Kunyu and Wang, Shaoan and Li, Minghan and Liu, Haoran and Wei, Songlin and Wang, Zhongyuan and Zhang, Zhizheng and Wang, He},
    journal={Robotics: Science and Systems},
    year={2025}
}

@inproceedings{liu2023llava,
    author      = {Liu, Haotian and Li, Chunyuan and Wu, Qingyang and Lee, Yong Jae},
    title       = {Visual Instruction Tuning},
    booktitle   = {NeurIPS},
    year        = {2023}
}

@article{kim2024openvla,
  title={Openvla: An open-source vision-language-action model},
  author={Kim, Moo Jin and Pertsch, Karl and Karamcheti, Siddharth and Xiao, Ted and Balakrishna, Ashwin and Nair, Suraj and Rafailov, Rafael and Foster, Ethan and Lam, Grace and Sanketi, Pannag and others},
  journal={arXiv preprint arXiv:2406.09246},
  year={2024}
}

@article{shen2024longvu,
  title={Longvu: Spatiotemporal adaptive compression for long video-language understanding},
  author={Shen, Xiaoqian and Xiong, Yunyang and Zhao, Changsheng and Wu, Lemeng and Chen, Jun and Zhu, Chenchen and Liu, Zechun and Xiao, Fanyi and Varadarajan, Balakrishnan and Bordes, Florian and others},
  journal={arXiv preprint arXiv:2410.17434},
  year={2024}
}

@inproceedings{roth2024viplanner,
  title={Viplanner: Visual semantic imperative learning for local navigation},
  author={Roth, Pascal and Nubert, Julian and Yang, Fan and Mittal, Mayank and Hutter, Marco},
  booktitle={2024 IEEE International Conference on Robotics and Automation (ICRA)},
  pages={5243--5249},
  year={2024},
  organization={IEEE}
}

@article{oquab2023dinov2,
  title={Dinov2: Learning robust visual features without supervision},
  author={Oquab, Maxime and Darcet, Timoth{\'e}e and Moutakanni, Th{\'e}o and Vo, Huy and Szafraniec, Marc and Khalidov, Vasil and Fernandez, Pierre and Haziza, Daniel and Massa, Francisco and El-Nouby, Alaaeldin and others},
  journal={arXiv preprint arXiv:2304.07193},
  year={2023}
}

@inproceedings{zhai2023siglip,
  title={Sigmoid loss for language image pre-training},
  author={Zhai, Xiaohua and Mustafa, Basil and Kolesnikov, Alexander and Beyer, Lucas},
  booktitle={Proceedings of the IEEE/CVF international conference on computer vision},
  pages={11975--11986},
  year={2023}
}

@article{qwen2,
    title   = {Qwen2 Technical Report}, 
    author  = {An Yang and Baosong Yang and Binyuan Hui and Bo Zheng and Bowen Yu and Chang Zhou and Chengpeng Li and Chengyuan Li and Dayiheng Liu and Fei Huang and Guanting Dong and Haoran Wei and Huan Lin and Jialong Tang and Jialin Wang and Jian Yang and Jianhong Tu and Jianwei Zhang and Jianxin Ma and Jin Xu and Jingren Zhou and Jinze Bai and Jinzheng He and Junyang Lin and Kai Dang and Keming Lu and Keqin Chen and Kexin Yang and Mei Li and Mingfeng Xue and Na Ni and Pei Zhang and Peng Wang and Ru Peng and Rui Men and Ruize Gao and Runji Lin and Shijie Wang and Shuai Bai and Sinan Tan and Tianhang Zhu and Tianhao Li and Tianyu Liu and Wenbin Ge and Xiaodong Deng and Xiaohuan Zhou and Xingzhang Ren and Xinyu Zhang and Xipin Wei and Xuancheng Ren and Yang Fan and Yang Yao and Yichang Zhang and Yu Wan and Yunfei Chu and Yuqiong Liu and Zeyu Cui and Zhenru Zhang and Zhihao Fan},
    journal = {arXiv preprint arXiv:2407.10671},
    year    = {2024}
}

@inproceedings{kostrikovoffline,
  title={Offline Reinforcement Learning with Implicit Q-Learning},
  author={Kostrikov, Ilya and Nair, Ashvin and Levine, Sergey},
  booktitle={International Conference on Learning Representations},
  year={2022}
}

@article{chen2025conrft,
  title={Conrft: A reinforced fine-tuning method for vla models via consistency policy},
  author={Chen, Yuhui and Tian, Shuai and Liu, Shugao and Zhou, Yingting and Li, Haoran and Zhao, Dongbin},
  journal={arXiv preprint arXiv:2502.05450},
  year={2025}
}

@article{liu2025can,
  title={What can rl bring to vla generalization? an empirical study},
  author={Liu, Jijia and Gao, Feng and Wei, Bingwen and Chen, Xinlei and Liao, Qingmin and Wu, Yi and Yu, Chao and Wang, Yu},
  journal={arXiv preprint arXiv:2505.19789},
  year={2025}
}

@article{huang2025co,
  title={CO-RFT: Efficient Fine-Tuning of Vision-Language-Action Models through Chunked Offline Reinforcement Learning},
  author={Huang, Dongchi and Fang, Zhirui and Zhang, Tianle and Li, Yihang and Zhao, Lin and Xia, Chunhe},
  journal={arXiv preprint arXiv:2508.02219},
  year={2025}
}

@article{zhang2025balancing,
  title={Balancing Signal and Variance: Adaptive Offline RL Post-Training for VLA Flow Models},
  author={Zhang, Hongyin and Zhang, Shiyuan and Jin, Junxi and Zeng, Qixin and Qiao, Yifan and Lu, Hongchao and Wang, Donglin},
  journal={arXiv preprint arXiv:2509.04063},
  year={2025}
}

@article{gao2025octonav,
  title={OctoNav: Towards Generalist Embodied Navigation},
  author={Gao, Chen and Jin, Liankai and Peng, Xingyu and Zhang, Jiazhao and Deng, Yue and Li, Annan and Wang, He and Liu, Si},
  journal={arXiv preprint arXiv:2506.09839},
  year={2025}
}

@article{qi2025vln,
  title={VLN-R1: Vision-Language Navigation via Reinforcement Fine-Tuning},
  author={Qi, Zhangyang and Zhang, Zhixiong and Yu, Yizhou and Wang, Jiaqi and Zhao, Hengshuang},
  journal={arXiv preprint arXiv:2506.17221},
  year={2025}
}

@article{li2025sekai,
      title={Sekai: A Video Dataset towards World Exploration}, 
      author={Zhen Li and Chuanhao Li and Xiaofeng Mao and Shaoheng Lin and Ming Li and Shitian Zhao and Zhaopan Xu and Xinyue Li and Yukang Feng and Jianwen Sun and Zizhen Li and Fanrui Zhang and Jiaxin Ai and Zhixiang Wang and Yuwei Wu and Tong He and Jiangmiao Pang and Yu Qiao and Yunde Jia and Kaipeng Zhang},
      journal={arXiv preprint arXiv:2506.15675},
      year={2025}
}

@article{savitzky1964smoothing,
  title={Smoothing and Differentiation of Data by Simplified Least Squares Procedures},
  author={Savitzky, Abraham and Golay, Marcel J. E.},
  journal={Analytical Chemistry},
  volume={36},
  number={8},
  pages={1627--1639},
  year={1964},
  publisher={ACS Publications},
  doi={10.1021/ac60214a047}
}

@article{van2011reciprocal,
  title={Reciprocal n-body collision avoidance},
  author={Van Den Berg, Jur and Guy, Stephen J and Lin, Ming and Manocha, Dinesh},
  journal={Robotics research},
  pages={3--19},
  year={2011},
  publisher={Springer}
}

@misc{schulman2017proximalpolicyoptimizationalgorithms,
      title={Proximal Policy Optimization Algorithms}, 
      author={John Schulman and Filip Wolski and Prafulla Dhariwal and Alec Radford and Oleg Klimov},
      year={2017},
      eprint={1707.06347},
      archivePrefix={arXiv},
      primaryClass={cs.LG},
}

@misc{fujimoto2019benchmarkingbatchdeepreinforcement,
      title={Benchmarking Batch Deep Reinforcement Learning Algorithms}, 
      author={Scott Fujimoto and Edoardo Conti and Mohammad Ghavamzadeh and Joelle Pineau},
      year={2019},
      eprint={1910.01708},
      archivePrefix={arXiv},
      primaryClass={cs.LG},
}

@misc{sun2021safeexplorationsolvingearly,
      title={Safe Exploration by Solving Early Terminated MDP}, 
      author={Hao Sun and Ziping Xu and Meng Fang and Zhenghao Peng and Jiadong Guo and Bo Dai and Bolei Zhou},
      year={2021},
      eprint={2107.04200},
      archivePrefix={arXiv},
      primaryClass={cs.LG},
}

@misc{fujimoto2021minimalistapproachofflinereinforcement,
      title={A Minimalist Approach to Offline Reinforcement Learning}, 
      author={Scott Fujimoto and Shixiang Shane Gu},
      year={2021},
      eprint={2106.06860},
      archivePrefix={arXiv},
      primaryClass={cs.LG},
}

@inproceedings{Bain1995AFF,
  title={A Framework for Behavioural Cloning},
  author={Michael Bain and Claude Sammut},
  booktitle={Machine Intelligence 15},
  year={1995},
}

@misc{ho2016generativeadversarialimitationlearning,
      title={Generative Adversarial Imitation Learning}, 
      author={Jonathan Ho and Stefano Ermon},
      year={2016},
      eprint={1606.03476},
      archivePrefix={arXiv},
      primaryClass={cs.LG},
}

@misc{zhang2025embodiednavigationfoundationmodel,
      title={Embodied Navigation Foundation Model}, 
      author={Jiazhao Zhang and Anqi Li and Yunpeng Qi and Minghan Li and Jiahang Liu and Shaoan Wang and Haoran Liu and Gengze Zhou and Yuze Wu and Xingxing Li and Yuxin Fan and Wenjun Li and Zhibo Chen and Fei Gao and Qi Wu and Zhizheng Zhang and He Wang},
      year={2025},
      eprint={2509.12129},
      archivePrefix={arXiv},
      primaryClass={cs.RO},
}

@article{abduljabbar2021role,
  title={The role of micro-mobility in shaping sustainable cities: A systematic literature review},
  author={Abduljabbar, Rusul L and Liyanage, Sohani and Dia, Hussein},
  journal={Transportation research part D: transport and environment},
  volume={92},
  pages={102734},
  year={2021},
  publisher={Elsevier}
}

@article{oeschger2020micromobility,
  title={Micromobility and public transport integration: The current state of knowledge},
  author={Oeschger, Giulia and Carroll, P{\'a}raic and Caulfield, Brian},
  journal={Transportation Research Part D: Transport and Environment},
  volume={89},
  pages={102628},
  year={2020},
  publisher={Elsevier}
}

@misc{amap2023,
  author       = {{AutoNavi}},
  title        = {Amap (Gaode) Maps},
  year         = {2023},
  howpublished = {[Online]. Available: \url{https://lbs.amap.com}},
  note         = {Accessed: Sep. 15, 2025}
}

@article{cadena2016slam,
  author    = {Cadena, Cesar and Carlone, Luca and Carrillo, Henry and Latif, Yasir and Scaramuzza, Davide and Neira, Jos{\'e} and Reid, Ian and Leonard, John J.},
  title     = {Past, present, and future of simultaneous localization and mapping: Toward the robust-perception age},
  journal   = {IEEE Transactions on Robotics},
  volume    = {32},
  number    = {6},
  pages     = {1309--1332},
  year      = {2016},
  doi       = {10.1109/TRO.2016.2624754}
}

@inproceedings{schlichting2016localization,
  author    = {Schlichting, Alexander and Brenner, Claus},
  title     = {Localization using automotive laser scanners and local pattern matching},
  booktitle = {ISPRS Annals of the Photogrammetry, Remote Sensing and Spatial Information Sciences},
  volume    = {3},
  number    = {1},
  pages     = {143--150},
  year      = {2016},
  doi       = {10.5194/isprsannals-III-1-143-2016}
}

@article{hata2016road,
  author    = {Hata, Alberto Y. and Wolf, Denis F.},
  title     = {Road marking detection using LIDAR reflective intensity data and its application to vehicle localization},
  journal   = {IEEE Intelligent Transportation Systems Magazine},
  volume    = {7},
  number    = {3},
  pages     = {18--29},
  year      = {2016},
  doi       = {10.1109/MITS.2015.2498190}
}

@article{wolcott2017lidar,
  author    = {Wolcott, Ryan W. and Eustice, Ryan M.},
  title     = {Robust LIDAR localization using multiresolution Gaussian mixture maps for autonomous driving},
  journal   = {The International Journal of Robotics Research},
  volume    = {36},
  number    = {3},
  pages     = {292--319},
  year      = {2017},
  doi       = {10.1177/0278364917691568}
}

@inproceedings{schreiber2013laneloc,
  author    = {Schreiber, Michael and Kn{\"o}ppel, Christian and Stiller, Christoph},
  title     = {LaneLoc: Lane marking based localization using highly accurate maps},
  booktitle = {Proc. IEEE Intelligent Vehicles Symposium (IV)},
  pages     = {449--454},
  year      = {2013},
  doi       = {10.1109/IVS.2013.6629518}
}

@inproceedings{zhang2014loam,
  author    = {Zhang, Ji and Singh, Sanjiv},
  title     = {LOAM: Lidar odometry and mapping in real-time},
  booktitle = {Proc. Robotics: Science and Systems (RSS)},
  year      = {2014},
  doi       = {10.15607/RSS.2014.X.007}
}

@article{Yang2023iPlannerIP,
  title={iPlanner: Imperative Path Planning},
  author={F. Yang and Chen Wang and C{\'e}sar Cadena and Marco Hutter},
  journal={ArXiv},
  year={2023},
  volume={abs/2302.11434}
}

@article{li2022metadrive,
  title={Metadrive: Composing diverse driving scenarios for generalizable reinforcement learning},
  author={Li, Quanyi and Peng, Zhenghao and Feng, Lan and Zhang, Qihang and Xue, Zhenghai and Zhou, Bolei},
  journal={IEEE transactions on pattern analysis and machine intelligence},
  volume={45},
  number={3},
  pages={3461--3475},
  year={2022},
  publisher={IEEE}
}

@article{deitke2022retrospectives,
  title={Retrospectives on the embodied ai workshop},
  author={Deitke, Matt and Batra, Dhruv and Bisk, Yonatan and Campari, Tommaso and Chang, Angel X and Chaplot, Devendra Singh and Chen, Changan and D'Arpino, Claudia P{\'e}rez and Ehsani, Kiana and Farhadi, Ali and others},
  journal={arXiv preprint arXiv:2210.06849},
  year={2022}
}

@misc{zhou2024mattersenhancetrafficrule,
      title={What Matters to Enhance Traffic Rule Compliance of Imitation Learning for End-to-End Autonomous Driving}, 
      author={Hongkuan Zhou and Wei Cao and Aifen Sui and Zhenshan Bing},
      year={2024},
      eprint={2309.07808},
      archivePrefix={arXiv},
      primaryClass={cs.CV},
}

@article{wu2025metaurban,
  title={MetaUrban: An Embodied AI Simulation Platform for Urban Micromobility},
  author={Wu, Wayne and He, Honglin and He, Jack and Wang, Yiran and Duan, Chenda and Liu, Zhizheng and Li, Quanyi and Zhou, Bolei},
  journal={ICLR},
  year={2025},
}

@misc{tong2024cambrian1,
      title={Cambrian-1: A Fully Open, Vision-Centric Exploration of Multimodal LLMs}, 
      author={Shengbang Tong and Ellis Brown and Penghao Wu and Sanghyun Woo and Manoj Middepogu and Sai Charitha Akula and Jihan Yang and Shusheng Yang and Adithya Iyer and Xichen Pan and Austin Wang and Rob Fergus and Yann LeCun and Saining Xie},
      year={2024},
      eprint={2406.16860},
}

@article{he2025from,
  title={From Seeing to Experiencing: Scaling Navigation Foundation Models with Reinforcement Learning},
  author={He, Honglin and Ma, Yukai and Wu, Wayne and Zhou, Bolei},
  journal={arXiv preprint arXiv:2507.22028},
  year={2025}
}

@ inproceedings{wu2025urbansim,
title={Towards Autonomous Micromobility through Scalable Urban Simulation},
author={Wu, Wayne and He, Honglin and Zhang, Chaoyuan and He, Jack and Zhao, Seth Z. and Gong, Ran and Li, Quanyi and Zhou, Bolei},
booktitle ={Proceedings of the IEEE/CVF Conference on Computer Vision and Pattern Recognition},
year={2025}
}

@misc{zhang2025crestescalablemaplessnavigation,
      title={CREStE: Scalable Mapless Navigation with Internet Scale Priors and Counterfactual Guidance}, 
      author={Arthur Zhang and Harshit Sikchi and Amy Zhang and Joydeep Biswas},
      year={2025},
      eprint={2503.03921},
      archivePrefix={arXiv},
      primaryClass={cs.RO},
}

@article{Wilde2014PushAR,
  title={Push and Rotate: a Complete Multi-agent Pathfinding Algorithm},
  author={Boris de Wilde and Adriaan ter Mors and Cees Witteveen},
  journal={J. Artif. Intell. Res.},
  year={2014},
  volume={51},
  pages={443-492},
}

@misc{kumar2020conservative,
      title={Conservative Q-Learning for Offline Reinforcement Learning}, 
      author={Aviral Kumar and Aurick Zhou and George Tucker and Sergey Levine},
      year={2020},
      eprint={2006.04779},
      archivePrefix={arXiv},
      primaryClass={cs.LG},
}

@misc{hirose2025omnivla,
      title={OmniVLA: An Omni-Modal Vision-Language-Action Model for Robot Navigation}, 
      author={Noriaki Hirose and Catherine Glossop and Dhruv Shah and Sergey Levine},
      year={2025},
      eprint={2509.19480},
      archivePrefix={arXiv},
      primaryClass={cs.RO},
}
